\documentclass{llncs}
\usepackage[pdftex]{graphicx}
\begin{document}

\title{Sample Dropout for Audio Scene Classification Using Multi-Scale Dense Connected Convolutional Neural Network}
\titlerunning{Multi-Scale DenseNet-Based Audio Scene Classification}  

\author{Dawei Feng\inst{1} \and Kele Xu\inst{1,2} \and
	Haibo Mi\inst{1} \and Feifan Liao\inst{2} \and Yan Zhou\inst{2}}
\authorrunning{Dawei Feng et al.} 
%
\tocauthor{Dawei Feng, Kele Xu, Haibo Mi, Feifan Liao and Yan Zhou}
\institute{Science and Technology on Parallel and Distributed Laboratory, School of Computer, National University of Defense Technology; Changsha, 410073, China\\
	\email{davyfeng.c@gmail.com}, \email{kelele.xu@gmail.com}, \email{haibo\_mihb@126.com}\\ 
	\and
	School of Information and Communication, National University of Defense Technology, Wuhan, 430010, China,\\
}

\maketitle 

\begin{abstract}
Acoustic scene classification is an intricate problem for a machine. As an emerging field of research, deep Convolutional Neural Networks (CNN) achieve convincing results. In this paper, we explore the use of multi-scale Dense connected convolutional neural network (DenseNet) for the classification task, with the goal to improve the classification performance as multi-scale features can be extracted from the time-frequency representation of the audio signal. On the other hand, most of previous CNN-based audio scene classification approaches aim to improve the classification accuracy, by employing different regularization techniques, such as the dropout of hidden units and data augmentation, to reduce overfitting. It is widely known that outliers in the training set have a high negative influence on the trained model, and culling the outliers may improve the classification performance, while it is often under-explored in previous studies. In this paper, inspired by the silence removal in the speech signal processing, a novel sample dropout approach is proposed, which aims to remove outliers in the training dataset. Using the DCASE 2017 audio scene classification datasets, the experimental results demonstrates the proposed multi-scale DenseNet providing a superior performance than the traditional single-scale DenseNet, while the sample dropout method can further improve the classification robustness of multi-scale DenseNet.
\keywords{sample dropout, audio scene classification, convolutional neural network, multi-scale}
\end{abstract}

\section{Introduction}
Acoustic scene classification (ASC) aims to distinguish between the various acoustical scenes, and effective identification of scenes through the analysis of unstructured patterns has many potential applications, such as, intelligent monitoring systems, context aware devices design and so on. Multiple overlapping sound sources are contained in the acoustic mark of a certain scene, as a result, ASC is of a great challenge despite the sustainable efforts have been made.

From the perspective of scene classification, different methods have been tested in the computer vision field, and dramatic progress has been made during last two decades, especially with the improvements of local invariant feature, such as (SIFT \cite{lowe2004distinctive}, SURF \cite{bay2006surf}) and convolutional neural network \cite{krizhevsky2012imagenet}. Nevertheless, compared to the image-based scene classification, audio-based approach is still under-explored. The state-of-the-art scene-classification methods using audio are not able to provide comparable accuracy with comparison to the image-based methods \cite{dixit2015scene}. However, the audio is more descriptive and salient than the images in some practical situations.

In the past several years, an increasing interest has been observed, which aims to find more robust and efficient approaches for acoustic scene classification and sound event detection, by using both supervised learning and unsupervised-learning methods. Specifically, the first Detection and Classification of Acoustic Scenes and Events (DCASE) 2013 challenge \cite{stowell2015detection} was organized by the IEEE Audio and Acoustic Signal Processing (AASP) Technical Committee, aims to solve the problem of lacking common benchmarking datasets, and has stimulated the research community to explore more efficient methods. Since the release of the relatively larger labeled data, there has been a plethora of efforts made for the audio scene classification task \cite{eghbal2016cp}, \cite{mesaros2016tut}, \cite{mesaros2017dcase}. 

Recently, deep learning have achieved convincing performance in different fields, ranging from computer vision \cite{krizhevsky2012imagenet}., speech recognition \cite{hinton2012deep} to natural language processing \cite{manning2014stanford}. Extensive deep learning architectures have been explored for the audio signal processing, for example, auto-encoder, convolutional neural network, recurrent neural network, and different regularization methods are also tested for the task. 

Most of the previous attempts aimed to apply the deep learning by modifying the CNN architectures. In this paper, we aim to improve the ASC performance by using the multi-scale DenseNet and culling sample-based regularization method. In more detail, multi-scale DenseNet is employed to extract multi-scale information embedded in the time-frequency of the audio signal. Moreover, unlike previous attempts to dropout hidden layers in the neural network training, we explore the low-variance sample dropout approaches, with the goal to culling the “outliers” in the training samples. After removing the specified samples in the training data set, the neural network classifier is trained with the remaining examples to obtain robust models. Using the DCASE 2017 audio scene classification dataset \cite{mesaros2017dcase}, our experimental evaluation shows that the proposed method can improve the robustness of the classifier.

The paper is organized as follows: Section 2 gives a short summary for the related work. Section 3 presents the data used and the experimental setup, while section 4 describes the multi-scale DenseNet. The approach to cull samples in the training set is given in Section 5. The experimental results are presented in section 6, while Section 7 concludes this paper.

\section{Related Work}

ASC is a complicated issue which aims at distinguishing acoustic environments solely depended on the audio recordings of the scene. 
During last decades, various feature extraction algorithms (representing the audio scenes), and classification models have been proposed in previous works. The most popular baseline is Gaussian Mixture Model (GMM) or Hidden Markov Model (HMM), by using the Mel-Frequency Cepstral Coefficients (MFCCs) \cite{mesaros2016tut}. Shallow-architecture classifier, such as, Support Vector Machines (SVM) \cite{geiger2013large} and Gradient Booting Trees (GBM) \cite{fonseca2017acoustic}, were also tested for the classification task. Moreover, non-negative matrix factorization (NMF) approach can be utilized to extract subspace representation prior to the classification.

Recently, many works demonstrated that deep neural network can improve the classification accuracy while no handcrafted features are needed. In brief, for the ASC task, the main modifications of deep learning-based approaches can be divided into three categories: deep learning using different representations of the audio signal \cite{aytar2016soundnet}, \cite{eghbal2016cp}; more sophisticated deep learning architectures classifiers \cite{marchi2016up}, \cite{bae2016acoustic} and the applications of different regularizations methods to train the deep neural network\cite{phan2017cnn}. 

Compared to the traditional methods, which commonly involved training a Gaussian Mixture Model (GMM) on the frame-level features \cite{mesaros2016tut}, deep learning-based methods can achieve better performance \cite{mesaros2017dcase}. The well-known deep neural network models include deep belief network (DBN) and auto-encoders, convolutional neural network (CNN), recurrent neural network (RNN). Here, CNN is selected as the classifier due to its high potential to identify the various patterns of audio signals. Moreover, unlike previous attempts, we explore the use of multi-scale DenseNet as additional features may be embedded in different time range. The combination of features from multi-scale may lead to more salient feature representations for the classification task.

On the other hand, the neural networks are vulnerable to the outliers in the training set, and the outliers in the training set have a high negative influence on the trained model. As a result, the pre-processing of training samples is a key factor for the audio scene classification accuracy \cite{xu2018mixup}, while it is often under-explored in previous studies. In this work, we explore to cull the train samples of low-variance, which can be viewed as the noise (or outliers), with the goal to improve the accuracy.

\section{Audio Scene Classification Datasets and Experimental Setup}

\begin{figure}
	\centering{\includegraphics[width=4in,height=5in,keepaspectratio]{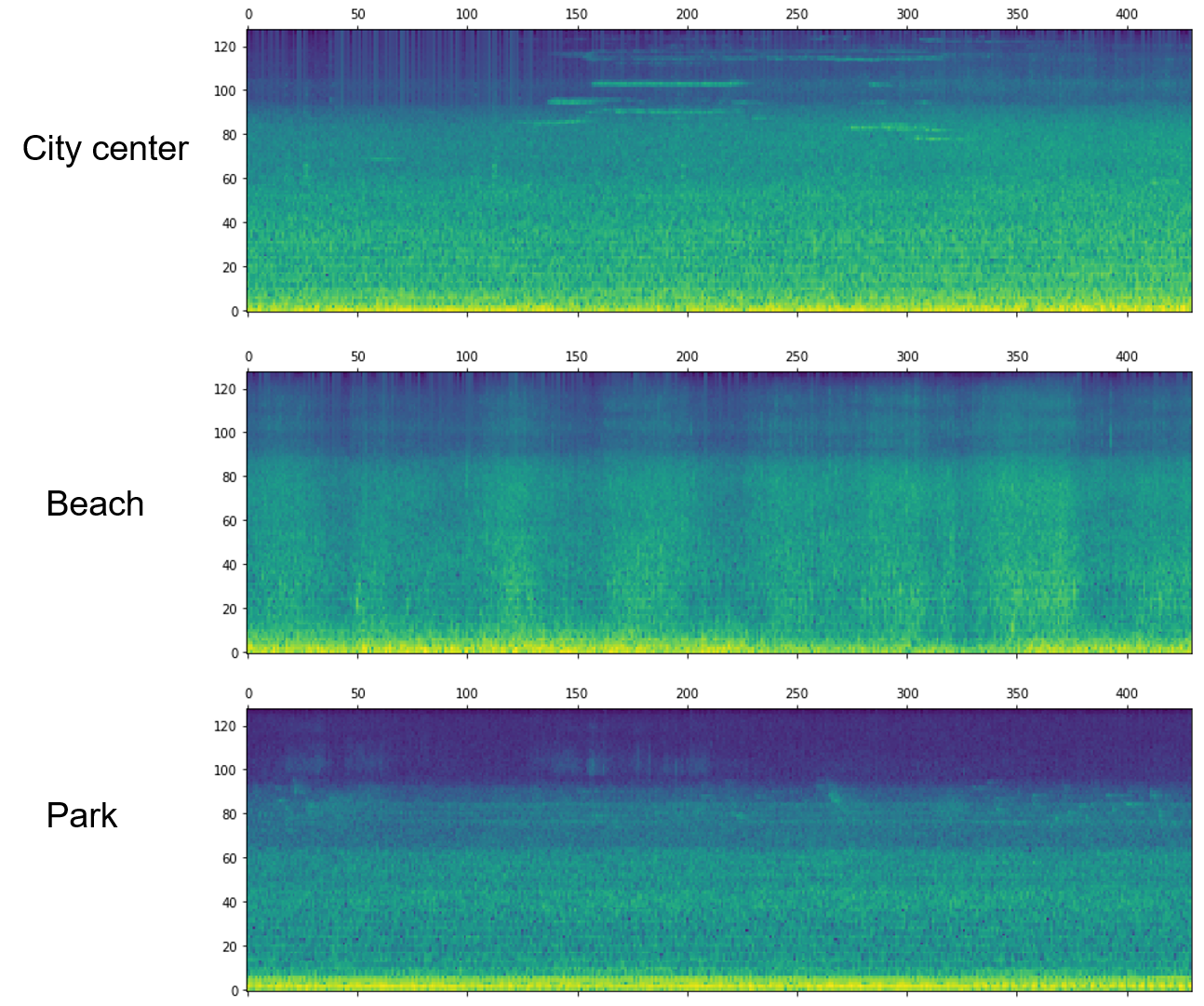}} 
	\caption{Mel filter bank energy of different audio scene.}
\end{figure}

As aforementioned, for the audio scene classification task, the well-collected data sets include: DCASE 2013 dataset \cite{stowell2015detection}, DCASE 2016 dataset \cite{mesaros2016tut}, DCASE 2017 dataset \cite{mesaros2017dcase}, Rouen dataset \cite{rakotomamonjy2015histogram} and ESL dataset \cite{piczak2015esc}.  We evaluate the proposed method on the TUT audio scene classification 2017 database \cite{mesaros2017dcase}, as this data is of large scale and covers 15 different acoustic scenes.

In more detail, the database consists of stereo recordings, which were collected using 44.1 kHz sampling rate and 24-bit resolution. The recordings came from various acoustic scenes, which have distinct recording locations. 3-5 minutes long audio was recorded for each sample, all of samples are divided into four cross-validation folds. And the audios were split into 10-second segments. The acoustic scene classes considered in this task were: bus, cafe/restaurant, car, city center, forest path, grocery store, home, lakeside beach, library, metro station, office, residential area, train, tram, and urban park.

As the input of the different deep neural network architectures, it can be either the raw audio signal or the time-frequency representation of the raw audio. Presently, most of the audio-related classification and detection system relied on the hand-crafted time-frequency representations of the audio signal. For example, Mel-frequency cepstrum (MFCCs) are widely used in speech recognition. However, MFCCs are developed inspired by the human speech production process, which assumes sounds are produced by glottal pulse passing through the vocal tract filter. MFCCs discard useful information about the sound, which restricts its ability for recognition and classification. In recent years, Mel Bank Features have been widely used in speaker recognition. In this paper, Mel-filter bank feature is used for our experiments, as mel filterbanks can provide better performance, and the mel-bank is reweighted to the same height (shown in Figure 1).

\section{Multi-Scale DenseNet}
Convolutional neural network (CNN) has a great potential to identify the various salient patterns of audio signals. There are numerous variants of CNN architectures in the literature. However, their basic components are very similar. Since the starting with LeNet-5 \cite{lecun1995learning}, convolutional neural networks have typically standard structure-stacked convolutional layers (optionally followed by batch normalization and max-pooling) are followed by fully-connected layers \cite{li2018multi}.

Many recent works claim that deeper CNN can provides better performance, as demonstrated by the progress on the image-classification task by using AlexNet \cite{krizhevsky2012imagenet}, VGGNet \cite{simonyan2014very}, ResNet \cite{he2016deep} architectures. Unlike traditional sequential network architectures such as AlexNet, ResNet is a form of architecture that relies on network-in-network block. DenseNet\cite{huang2017densely} is a new architecture and it is a logical extension of ResNet.

In more detail, DenseNet has a fundamental building block, which connects each layer to every other layer in a dense manner. There is a direct connection between any two layers in each dense block, and the input for each layer is the union of the outputs from all the previous layers. Compared to the conventional CNN, DenseNet not only performs better in image classification, but also provide a higher utilization rate for the original data and less feature information loss. It reinforces feature propagation, supports feature re-utilization, solves vanishing gradient problems effectively and significantly reduce the number of parameters.

For the audio scene classification task, the input for the DenseNet can the time-frequency representation of the raw audio signal. In this paper, as aforementioned, the Mel filter-bank energy is selected as the input of neural network. However, it is widely known that the multi-scale information are embedded in the time-frequency representation of the audio signal. Thus, automatically fully extraction of the multi-scale features is of great importance to improve the classification accuracy. A multi-scale DenseNet \cite{huang2017multi} architecture is given in Figure 2.

\begin{figure}
	\centering{\includegraphics[width=4.5in,height=8in,keepaspectratio]{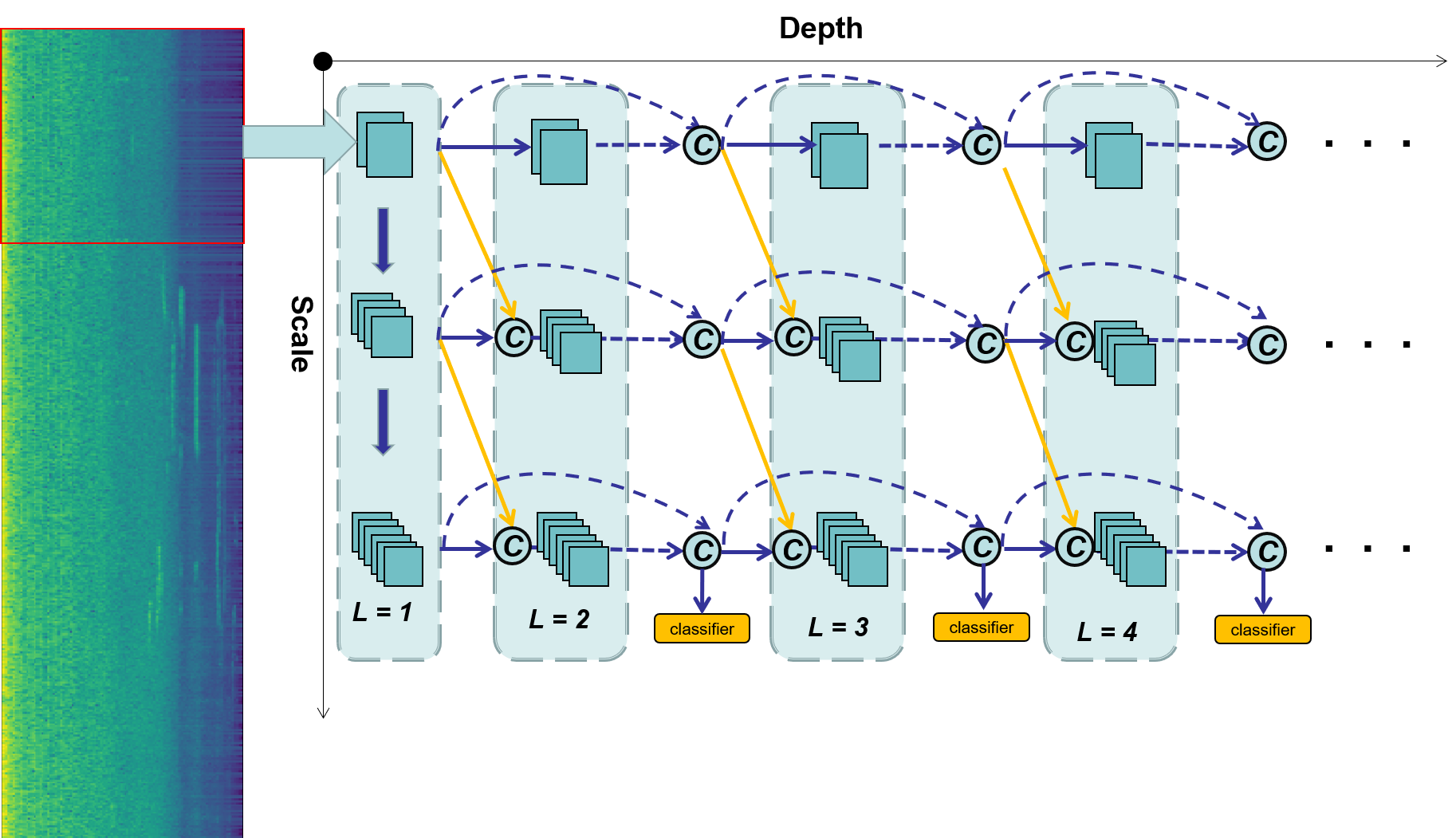}} 
	\caption{The multi-scale DenseNet architecture employed for audio scene classification task.}
\end{figure}

As shown in Figure 2, the proposed system architecture comprises of dense convolutional blocks with direct connections from any layer to all subsequent layers to improve the information flow on 128 $\times$ 128 input images. Four dense blocks with unequal numbers of layers make up the DenseNet used in our experiments.

A convolution with 64 output channels is performed on the input images in front of the first dense block. For convolutional layers with kernel size 3$\times$3, one-pixel padding is applied at each side of the inputs to keep the feature-map size fixed. The layers between two contiguous dense blocks are referred as transition layers for convolution and pooling, which contain 1$\times$1 convolution and 2$\times$2 average pooling. A bottleneck layer is used by using a 1$\times$1 convolution before each 3$\times$3 convolution in order to reduce the number of input feature-maps and improve computational efficiency. At the end of the last dense block, a global average pooling and a softmax classifier are employed.

The multi-scale dense block is employed to capture the multi-scale features embedded within the Mel filter bank features. The length of the convolution kernel is longer in the upper layer and can generate more feature maps to extract long-term and medium-term features in the Mel filter bank energy representation of the audio signal. In the lower layer, the convolution kernel receives all outputs in the previous layers and the original input. As a result, it allows us to decrease the length of the convolution kernel and prompt the feature maps to grasp more short-term features. Since the network structure supports to re-utilize the feature maps, the multi-scale dense block concatenates all periodic features with different lengths into the transition layer for further manipulations. After this block, the model can extract a certain number of multi-scale features.

\section{Culling training samples for convolutional neural network}
It is widely known that convolutional neural networks have a natural tendency towards overfitting, especially when the training data size is small. DCASE 2017 audio scene classification dataset have much larger data size with comparison to the datasets used in the previous studies. However, compared the data size to train a neural network for the image classification task, the size is still small. Dropout and data augmentation are proved to be effective regularization approaches to train the convolutional neural network. In more detail,  dropout of neural in the hidden layer is widely used in a plethora of literatures, and the activation of every hidden unit is random randomly removed with a predefined probability \cite{srivastava2014dropout}. While, another common method to reduce overfitting on image data is to artificially enlarge the dataset using label-preserving transformations for data augmentation. However, artificially enlarge the dataset may induce more ambiguous data (which can be considered as the outliers) on which the model has low confidence and thus high loss. In the audio scene classification task, it is widely known that the convolutional neural networks are vulnerable to the ambiguous samples.

In this paper, unlike previous attempts, we aim to reduce the training set by removing outliers. In more detail, we tested two different kinds of methods. First, inspired by the silence remove employed for the speech recognition, we remove the training samples, which associated with low-magnitude of the audio signal segment. It is well-known that silence-remove can boost the performance of the automatic speech recognition.
 
Secondly, we explore to cull low-variance training samples directly. As can be seen from the Figure 1, flatten regions exits in the mel-bank features of the audio signal. However, these flatten regions may contain few useful information during the network training as pixel values are almost the same in the whole sample.  Our insight is that by removing the low-variance samples, we can make our model to be more robust without changing either the network architecture or training method. 

In our experiments, we trained the classifier by using the multi-scale DenseNet to test every example in the training set, and considered examples whose variances were below a threshold as outliers, which will be removed during the majority voting.

\section{Experimental results}
 
In our experimental stage, the average accuracy is calculated by using pre-defined four folds for the cross-validation. For comparison, the baselines are also given in our experiments. The baseline includes a Gaussian mixture model (GMM) (with 16 Gassians per class).

In more detail, the baseline system used here consists of 60 MFCC features and a Gaussian mixture model based classifier. MFCCs were calculated using 40-ms frames with Hamming window and 50\% overlap and 40 mel bands. They include the first 20 coefficients (including the 0th order coefficient) and delta and acceleration coefficients calculated using a window length of 9 frames. A GMM model with 32 components was trained for each scene class. To train the multi-scale DenseNet, we used Keras with tensorflow backend, which can fully utilize GPU resource. CUDA and cuDNN were also used to accelerate the system.

The audio scene classification accuracies obtained by different dropout ratio are given in Table 1. During our experiments, the silence remove-based sample dropout method does not boost the performance, thus the culling samples of low-variance, using mel filterbank energy representation, is reported in our experiments. Due to the page limitation, only the multi-scale DenseNet-based audio scene classification accuracy is reported, by using different sample dropout ratio. As can be seen from the table, the sample dropout can improve the performance for the classification task. While, 0.01 percentage of low-variance samples dropout processing provides superior performance. The dropout ratio 0.01 is employed in our following experiments.

\begin{table}[htbp]
	\caption{Multi-scale DenseNet-based audio scene classification accuracy using different sample dropout ratio}
	\begin{center}
		\begin{tabular}{|c|c|c|}
			\cline{1-3} 
			\textbf{\textit{Sample dropout ratio}}& \textbf{\textit{Cross-validation}}& \textbf{\textit{Evaluation}} \\
			\hline
			0 & 80.3\%  & 70.6\%  \\
			0.01  & 83.4\%  & 72.5\%  \\
			0.1   & 82.5\%  & 71.8\%  \\
			0.2   & 80.2\%  & 70.7\%  \\
			\hline
		\end{tabular}
		\label{tab1}
	\end{center}
\end{table}

Table 2 provides a further quantitative comparison between different methods. In our experiments, both original DenseNet and multi-scale DenseNet showed better performance against the baseline. Without any data augmentation, the original cross-validation accuracy of multi-scale DenseNet is 80.4\%, while the original (single scale) DenseNet provides the 78.3\% accuracy. With sample dropout, the performance of original DenseNet and multi-scale DenseNet are boosted. The performance of baseline is also improved with sample dropout.

For the performance on the evaluation dataset, the accuracy of baseline was improved from 61.0\% to 63.4\% with the sample dropout. The accuracy of original DenseNet was improved from 68.8\% to 69.5\% with the sample dropout, while the accuracy of multi-scale DenseNet was improved from 68.8\% to 69.5\% with the sample dropout.

\begin{table}[htbp]
	\caption{Audio scene classification accuracy using single/multi-scale DenseNet and sample dropout (0.01 ratio)}
	\begin{center}
		\begin{tabular}{|c|c|c|}
			\cline{1-3} 
			\textbf{\textit{Method}}& \textbf{\textit{Cross-validation}}& \textbf{\textit{Evaluation}} \\
			\hline
			Baseline (GMM) & 74.8\%  & 61.0\%  \\
			DenseNet & 78.3\%  & 68.8\%  \\
			Multi-scale using DenseNet & 80.4\%  & 70.5\%  \\
			\hline
			Baseline (GMM) with sample dropout & 76.2\%  & 63.4\%  \\
			DenseNet with sample dropout & 80.6\%  & 69.5\%  \\
			Multi-scale using DenseNet with sample dropout & 83.4\%  & 72.5\%  \\
			\hline
		\end{tabular}
		\label{tab1}
	\end{center}
\end{table}

As can be seen from the table, the results demonstrates that multi-scale DenseNet showing a better performance than the original single-scale DenseNet. It may imply that additional features can be extracted from multi-scale, which can improve the accuracy for the audio scene classification task. Moreover, the sample dropout method can boost the performance further, which demonstrates the effectiveness of proposed method.

\section{Conclusion}
In this paper, we have presented the multi-scale DenseNet-based method for the multi-class acoustic scene classification.
To summarize, the contributions of this paper are twofold: firstly, we explore a multi-scale CNN architecture for the classification task. To the best knowledge of the authors, this is the first attempt to employ multi-scale DenseNet for the audio scene classification task. Secondly, we propose a novel sample dropout method, and experiments demonstrate that by employing the proposed sample dropout approach, the classification performance can be improved further.
For future work, we will conduct a quantitative comparison between different widely-used CNN architectures, which can be helpful to design specified architecture for the audio scene classification task. Moreover, the transfer learning-based audio scene classification also needs to be explored in future work.

\section*{Acknowledgement}
This study was supported by the Strategic Priority Research Programme(17-ZLXD-XX-02-06-02-08). We also thank for Qiuqiang Kong from Centre for Vision, Speech and Signal Processing (CVSSP), Surrey University, for providing helpful suggestions.

%
%
\bibliographystyle{IEEEtran}

\bibliography{ref_collection_kele}

\begin{thebibliography}{10}
\providecommand{\url}[1]{#1}
\csname url@samestyle\endcsname
\providecommand{\newblock}{\relax}
\providecommand{\bibinfo}[2]{#2}
\providecommand{\BIBentrySTDinterwordspacing}{\spaceskip=0pt\relax}
\providecommand{\BIBentryALTinterwordstretchfactor}{4}
\providecommand{\BIBentryALTinterwordspacing}{\spaceskip=\fontdimen2\font plus
\BIBentryALTinterwordstretchfactor\fontdimen3\font minus
  \fontdimen4\font\relax}
\providecommand{\BIBforeignlanguage}[2]{{%
\expandafter\ifx\csname l@#1\endcsname\relax
\typeout{** WARNING: IEEEtran.bst: No hyphenation pattern has been}%
\typeout{** loaded for the language `#1'. Using the pattern for}%
\typeout{** the default language instead.}%
\else
\language=\csname l@#1\endcsname
\fi
#2}}
\providecommand{\BIBdecl}{\relax}
\BIBdecl

\bibitem{lowe2004distinctive}
D.~G. Lowe, ``Distinctive image features from scale-invariant keypoints,''
  \emph{International journal of computer vision}, vol.~60, no.~2, pp. 91--110,
  2004.

\bibitem{bay2006surf}
H.~Bay, T.~Tuytelaars, and L.~Van~Gool, ``Surf: Speeded up robust features,''
  in \emph{European conference on computer vision}.\hskip 1em plus 0.5em minus
  0.4em\relax Springer, 2006, pp. 404--417.

\bibitem{krizhevsky2012imagenet}
A.~Krizhevsky, I.~Sutskever, and G.~E. Hinton, ``Imagenet classification with
  deep convolutional neural networks,'' in \emph{Advances in neural information
  processing systems}, 2012, pp. 1097--1105.

\bibitem{dixit2015scene}
M.~Dixit, S.~Chen, D.~Gao, N.~Rasiwasia, and N.~Vasconcelos, ``Scene
  classification with semantic fisher vectors,'' in \emph{Computer Vision and
  Pattern Recognition (CVPR), 2015 IEEE Conference on}.\hskip 1em plus 0.5em
  minus 0.4em\relax IEEE, 2015, pp. 2974--2983.

\bibitem{stowell2015detection}
D.~Stowell, D.~Giannoulis, E.~Benetos, M.~Lagrange, and M.~D. Plumbley,
  ``Detection and classification of acoustic scenes and events,'' \emph{IEEE
  Transactions on Multimedia}, vol.~17, no.~10, pp. 1733--1746, 2015.

\bibitem{eghbal2016cp}
H.~Eghbal-Zadeh, B.~Lehner, M.~Dorfer, and G.~Widmer, ``Cp-jku submissions for
  dcase-2016: A hybrid approach using binaural i-vectors and deep convolutional
  neural networks,'' \emph{IEEE AASP Challenge on Detection and Classification
  of Acoustic Scenes and Events (DCASE)}, 2016.

\bibitem{mesaros2016tut}
A.~Mesaros, T.~Heittola, and T.~Virtanen, ``Tut database for acoustic scene
  classification and sound event detection,'' in \emph{Signal Processing
  Conference (EUSIPCO), 2016 24th European}.\hskip 1em plus 0.5em minus
  0.4em\relax IEEE, 2016, pp. 1128--1132.

\bibitem{mesaros2017dcase}
A.~Mesaros, T.~Heittola, A.~Diment, B.~Elizalde, A.~Shah, E.~Vincent, B.~Raj,
  and T.~Virtanen, ``Dcase 2017 challenge setup: Tasks, datasets and baseline
  system,'' in \emph{DCASE 2017-Workshop on Detection and Classification of
  Acoustic Scenes and Events}, 2017.

\bibitem{hinton2012deep}
G.~Hinton, L.~Deng, D.~Yu, G.~E. Dahl, A.-r. Mohamed, N.~Jaitly, A.~Senior,
  V.~Vanhoucke, P.~Nguyen, T.~N. Sainath \emph{et~al.}, ``Deep neural networks
  for acoustic modeling in speech recognition: The shared views of four
  research groups,'' \emph{IEEE Signal Processing Magazine}, vol.~29, no.~6,
  pp. 82--97, 2012.

\bibitem{manning2014stanford}
C.~Manning, M.~Surdeanu, J.~Bauer, J.~Finkel, S.~Bethard, and D.~McClosky,
  ``The stanford corenlp natural language processing toolkit,'' in
  \emph{Proceedings of 52nd annual meeting of the association for computational
  linguistics: system demonstrations}, 2014, pp. 55--60.

\bibitem{geiger2013large}
J.~T. Geiger, B.~Schuller, and G.~Rigoll, ``Large-scale audio feature
  extraction and svm for acoustic scene classification,'' in \emph{Applications
  of Signal Processing to Audio and Acoustics (WASPAA), 2013 IEEE Workshop
  on}.\hskip 1em plus 0.5em minus 0.4em\relax IEEE, 2013, pp. 1--4.

\bibitem{fonseca2017acoustic}
E.~Fonseca, R.~Gong, D.~Bogdanov, O.~Slizovskaia, E.~G{\'o}mez~Guti{\'e}rrez,
  and X.~Serra, ``Acoustic scene classification by ensembling gradient boosting
  machine and convolutional neural networks,'' in \emph{Virtanen T, Mesaros A,
  Heittola T, Diment A, Vincent E, Benetos E, Martinez B, editors. Detection
  and Classification of Acoustic Scenes and Events 2017 Workshop (DCASE2017);
  2017 Nov 16; Munich, Germany. Tampere (Finland): Tampere University of
  Technology; 2017. p. 37-41.}\hskip 1em plus 0.5em minus 0.4em\relax Tampere
  University of Technology, 2017.

\bibitem{aytar2016soundnet}
Y.~Aytar, C.~Vondrick, and A.~Torralba, ``Soundnet: Learning sound
  representations from unlabeled video,'' in \emph{Advances in Neural
  Information Processing Systems}, 2016, pp. 892--900.

\bibitem{marchi2016up}
E.~Marchi, D.~Tonelli, X.~Xu, F.~Ringeval, J.~Deng, and B.~Schuller, ``The up
  system for the 2016 dcase challenge using deep recurrent neural network and
  multiscale kernel subspace learning,'' \emph{Detection and Classification of
  Acoustic Scenes and Events}, 2016.

\bibitem{bae2016acoustic}
S.~H. Bae, I.~Choi, and N.~S. Kim, ``Acoustic scene classification using
  parallel combination of lstm and cnn,'' in \emph{Proceedings of the Detection
  and Classification of Acoustic Scenes and Events 2016 Workshop (DCASE2016)},
  2016, pp. 11--15.

\bibitem{phan2017cnn}
H.~Phan, P.~Koch, L.~Hertel, M.~Maass, R.~Mazur, and A.~Mertins, ``Cnn-lte: a
  class of 1-x pooling convolutional neural networks on label tree embeddings
  for audio scene classification,'' in \emph{Acoustics, Speech and Signal
  Processing (ICASSP), 2017 IEEE International Conference on}.\hskip 1em plus
  0.5em minus 0.4em\relax IEEE, 2017, pp. 136--140.

\bibitem{xu2018mixup}
K.~Xu, D.~Feng, H.~Mi, B.~Zhu, D.~Wang, L.~Zhang, H.~Cai, and S.~Liu,
  ``Mixup-based acoustic scene classification using multi-channel convolutional
  neural network,'' \emph{arXiv preprint arXiv:1805.07319}, 2018.

\bibitem{rakotomamonjy2015histogram}
A.~Rakotomamonjy and G.~Gasso, ``Histogram of gradients of time-frequency
  representations for audio scene classification,'' \emph{IEEE/ACM Transactions
  on Audio, Speech and Language Processing (TASLP)}, vol.~23, no.~1, pp.
  142--153, 2015.

\bibitem{piczak2015esc}
K.~J. Piczak, ``Esc: Dataset for environmental sound classification,'' in
  \emph{Proceedings of the 23rd ACM international conference on
  Multimedia}.\hskip 1em plus 0.5em minus 0.4em\relax ACM, 2015, pp.
  1015--1018.

\bibitem{lecun1995learning}
Y.~LeCun, L.~Jackel, L.~Bottou, C.~Cortes, J.~S. Denker, H.~Drucker, I.~Guyon,
  U.~Muller, E.~Sackinger, P.~Simard \emph{et~al.}, ``Learning algorithms for
  classification: A comparison on handwritten digit recognition,'' \emph{Neural
  networks: the statistical mechanics perspective}, vol. 261, p. 276, 1995.

\bibitem{li2018multi}
B.~Li, K.~Xu, X.~Cui, Y.~Wang, X.~Ai, and Y.~Wang, ``Multi-scale densenet-based
  electricity theft detection,'' \emph{arXiv preprint arXiv:1805.09591}, 2018.

\bibitem{simonyan2014very}
K.~Simonyan and A.~Zisserman, ``Very deep convolutional networks for
  large-scale image recognition,'' \emph{arXiv preprint arXiv:1409.1556}, 2014.

\bibitem{he2016deep}
K.~He, X.~Zhang, S.~Ren, and J.~Sun, ``Deep residual learning for image
  recognition,'' in \emph{Proceedings of the IEEE conference on computer vision
  and pattern recognition}, 2016, pp. 770--778.

\bibitem{huang2017densely}
G.~Huang, Z.~Liu, K.~Q. Weinberger, and L.~van~der Maaten, ``Densely connected
  convolutional networks,'' in \emph{Proceedings of the IEEE conference on
  computer vision and pattern recognition}, vol.~1, no.~2, 2017, p.~3.

\bibitem{huang2017multi}
G.~Huang, D.~Chen, T.~Li, F.~Wu, L.~van~der Maaten, and K.~Q. Weinberger,
  ``Multi-scale dense convolutional networks for efficient prediction,''
  \emph{arXiv preprint arXiv:1703.09844}, 2017.

\bibitem{srivastava2014dropout}
N.~Srivastava, G.~Hinton, A.~Krizhevsky, I.~Sutskever, and R.~Salakhutdinov,
  ``Dropout: A simple way to prevent neural networks from overfitting,''
  \emph{The Journal of Machine Learning Research}, vol.~15, no.~1, pp.
  1929--1958, 2014.

\end{thebibliography}

\end{document}